\documentclass[10pt,twocolumn,letterpaper]{article}

\usepackage{cvpr}              

\usepackage[dvipsnames]{xcolor}

\usepackage{booktabs}
\usepackage{multirow}
\usepackage{multicol}
\usepackage{colortbl}
\usepackage{amsmath}

\definecolor{cvprblue}{rgb}{0.21,0.49,0.74}
\usepackage[pagebackref,breaklinks,colorlinks,citecolor=cvprblue]{hyperref}

\newcommand{\name}{Efficient Grounded Prompting and Adaptation}
\newcommand{\namebf}{\textbf{Gro}unded \textbf{Prompt}ing}
\newcommand{\nameshort}{GroPrompt}

\newcommand{\namea}{Text-Contrastive Prompt Learning}

\newcommand{\nameaabbr}{TextCon}
\newcommand{\nameb}{Modality-Contrastive Prompt Learning}

\newcommand{\namebabbr}{ModalCon}
\newcommand{\namef}{Text-Aware Prompt Contrastive Learning}
\newcommand{\namefbf}{\textbf{T}ext-\textbf{A}ware \textbf{P}rompt \textbf{C}ontrastive \textbf{L}earning}
\newcommand{\namefabbr}{TAP-CL}

\def\paperID{19} 
\def\confName{CVPR}
\def\confYear{2024}

\title{\nameshort: \name~ \\for Referring Video Object Segmentation}

\author{Ci-Siang Lin$^{1,2}$$^*$$^\dagger$ \and I-Jieh Liu$^{1}$$^*$ \and Min-Hung Chen$^{2}$ \and Chien-Yi Wang$^{2}$ \and Sifei Liu$^{2}$ \and Yu-Chiang Frank Wang$^{1, 2}$ \\
$^{1}$Graduate Institute of Communication Engineering, National Taiwan University, Taiwan \\ $^{2}$NVIDIA\\
{\tt\small \{d08942011, r11942087, ycwang\}@ntu.edu.tw, \{minghungc, chienyiw, sifeil\}@nvidia.com}
}

\begin{document}
\maketitle

\def\thefootnote{*}\footnotetext{Equal contribution.}

\def\thefootnote{$\dagger$}\footnotetext{Work done during an internship at NVIDIA.}

\begin{abstract}

Referring Video Object Segmentation (RVOS) aims to segment the object referred to by the query sentence throughout the entire video. 
Most existing methods require end-to-end training with dense mask annotations, which could be computation-consuming and less scalable. 
In this work, we aim to efficiently adapt foundation segmentation models for addressing RVOS from weak supervision with the proposed \namebf~(\textbf{\nameshort}) framework. 
More specifically, we propose \textit{\namefbf~(\textbf{\namefabbr})} to enhance the association between the position prompts and the referring sentences with only box supervisions, including \textit{\namea~(\nameaabbr)}~and \textit{\nameb~(\namebabbr)} at frame level and video level, respectively. 
With the proposed \namefabbr, our \nameshort~framework can generate temporal-consistent yet text-aware position prompts describing locations and movements for the referred object from the video. 
The experimental results in the standard RVOS benchmarks (Ref-YouTube-VOS, Ref-DAVIS17, A2D-Sentences, and JHMDB-Sentences) demonstrate the competitive performance of our proposed \nameshort~framework given only bounding box weak supervisions.

\end{abstract}
\section{Introduction}
\label{sec:intro}

Referring Video Object Segmentation (RVOS), aims to segment the object referred to by a sentence query throughout the entire video. In contrast to RIS, RVOS is particularly faced with dynamic visual challenges, such as position and size variation, pose deformation, object occlusion or exit, and scene variation. Moreover, the referring sentence may contain long-term motions or actions (\eg, ``a gold fish on the left swimming towards the top right''), which could not be easily recognized from a single frame. To address this challenging task, many works~\cite{seo2020urvos,wu2022language,han2023html,wu2023onlinerefer,miao2023spectrum,tang2023temporal,luo2023soc,wu2023segment,li2023learning} have been proposed. URVOS~\cite{seo2020urvos} is pioneering as a unified framework for referring video segmentation, which introduces memory attention modules to retrieve relevant information from the previous frame and encourage temporal consistency. With the rapid development of Transformer, ReferFormer~\cite{wu2022language} adopts encoder and decoder layers in the Transformer model and views language as queries to attend to the referred object, and an instance matching strategy is utilized to achieve object tracking. Recent works like FS-RVOS~\cite{li2023learning} and OnlineRefer~\cite{wu2023onlinerefer} further extend RVOS into the few-shot setting and online pipeline to handle limited samples and ongoing videos in real-world scenarios, respectively. Nevertheless, most existing methods require end-to-end training for vision-language models, which could be computationally expensive and time-consuming. Moreover, the requirement of dense mask annotations for training impedes the scalability of those approaches.

Recently, foundation segmentation models~\cite{kirillov2023segment,wang2023seggpt,zou2023segment} has been proposed. By leveraging numerous training data and employing large-scale model architectures, they can produce high-quality object masks according to various prompts such as points or boxes, and have shown overwhelming generalizability on various datasets, setting superior benchmarks for segmentation tasks. 
However, there are still challenges in the RVOS problem not addressed by those foundation models. For example, 
SAM~\cite{kirillov2023segment} is trained solely with images and their associated masks, not tailored to handle natural language descriptions and video data in RVOS. While it is possible to adapt SAM to the task of RVOS by incorporating grounding models (\eg,~\cite{liu2023grounding}) to generate text-associated position prompts and tracking models (\eg,~\cite{cheng2023segment}) to capture object motions across video frames, such naive combination of off-the-shelf models has shown to be suboptimal~\cite{li2023refsam}, as they are individually trained for different tasks. 
Therefore, a question arises: ``\textit{How can we effectively exploit foundation segmentation models to address RVOS?}''
We argue that the RVOS problem can be decomposed into \textit{referring}, \textit{video}, and \textit{segmentation} factors, and leave the segmentation problem to foundation segmentation models. We only focus on addressing the referring and video factors as current foundation models can already tackle to segmentation problem effectively. 


In this paper, we aim to efficiently adapt image-based foundation segmentation models for addressing referring video object segmentation from weak supervision. To achieve this goal, we propose a novel \textit{\namebf~(\textbf{\nameshort})} framework, which advances vision-language learning to produce temporal-consistent yet text-aware position prompts for segmentation purposes. 
More specifically, we propose \textit{\namefbf~(\textbf{\namefabbr})} to enhance the association between the position prompts and the referring sentences with only box supervisions, including \textit{\namea~(\nameaabbr)}~and \textit{\nameb~(\namebabbr)} at frame level and video level, respectively. 
For \nameaabbr, we enforce our \nameshort~framework to generate distinct position prompts for different referring sentences within each video frame. As for the \namebabbr, given that the sentence description may contain long-term motions or actions spanning across different moments, we propose to align the whole sequence of position prompts and the corresponding object with the input text for each video clip. 
With the proposed \namefabbr, our \nameshort~framework can generate temporal-consistent yet text-aware position prompts describing locations and movements for the referred object from the video. More importantly, our derived position prompts would be utilized to instruct image-based foundation segmentation models to produce object masks, enabling efficient adaptation to referring video object segmentation without requiring dense mask annotations. 
The experimental results in the standard RVOS benchmarks (Ref-YouTube-VOS, Ref-DAVIS17, A2D-Sentences, and JHMDB-Sentences) demonstrate the competitive performance of our proposed \nameshort~framework given only bounding box weak supervisions.

 We highlight the contributions of this paper as follows:
 \begin{itemize}

    \item {We propose a novel \textit{\namebf~(\textbf{\nameshort})} framework, which performs efficient prompting and adapts image-based segmentation models to address referring video object segmentation without additional finetuning.\\}
    \item {To generate temporal-consistent yet text-aware position prompts for segmentation purposes, we propose to jointly perform \textit{\namea}~and \textit{\nameb} at frame-level and video-level, respectively.\\}
    \item {The derived position prompts would be utilized to instruct image-based foundation segmentation models to produce object masks, enabling efficient adaptation to referring video object segmentation with $7\times$ fewer trainable parameters compared with SOTAs.}
    
\end{itemize}
\section{Related Work}
\label{sec:related_work}

\subsection{Referring Video Object Segmentation}

Referring Video Object Segmentation (RVOS)~\cite{wu2022language,han2023html,wu2023onlinerefer,miao2023spectrum,tang2023temporal,luo2023soc,wu2023segment} strives to segment the object described by a free-form sentence query across the entire video duration. 
Recently, ReferFormer~\cite{wu2022language} views language as queries to pay attention to the referred object by adopting an encoder-decoder style in the transformer. However, this work only supports offline training and inference, limiting its usage in real-world scenarios. More recently, 
OnlineRefer~\cite{wu2023onlinerefer} further proposes an online RVOS setting to deal with the issues about offline limits, which makes it more possible to adapt to real-world scenarios. 
Nevertheless, most existing methods require end-to-end training for vision-language models, which could be computationally expensive and time-consuming. Moreover, the requirement of dense mask annotations for training impedes the scalability of those approaches.
Instead, we propose to exploit foundation segmentation models without text- and temporal-aware prompting, which is trained without mask annotations and supports online settings.

\subsection{Foundation Segmentation Models}
In recent years, foundation vision models have gained massive attention given their remarkable generalization capabilities on various downstream tasks. More recently, SAM~\cite{kirillov2023segment} has introduced a foundation model specifically tailored for segmentation tasks. SAM allows specific position prompts (\eg, points, boxes, \etc.) to demonstrate the zero-shot ability on the open vocabulary segmentation tasks with novel image distributions. Several works have studied the versatility of SAM, including remote sensing images~\cite{chen2023rsprompter, SAMRS}, medical image analysis~\cite{MedSAM, chen2023sam, wu2023medical, cheng2023sammed2d}, and adaptation to video-based tracking task~\cite{cheng2023segment, yang2023track, sam-pt}, \etc. 


For adaptation to tracking tasks with SAM, SAM-PT~\cite{sam-pt} 
designs a point-based prompt enhancement for the original SAM point prompt to support classic video object segmentation tasks, while neglecting the importance of text prompt for advanced referring video object segmentation. Another example SAM-Track~\cite{cheng2023segment} attempts to utilize SAM for segmentation and detection of objects while the DeAOT~\cite{yang2022decoupling} module captures the motion across frames for tracking the objects. Though it is possible to combine text-grounding detection models (\eg, Grounding DINO~\cite{liu2023grounding}) with SAM-Track to tackle RVOS, RefSAM~\cite{li2023refsam} has studied the possible concerns and indicates the unsatisfactory performance compared with current SOTAs in RVOS tasks. 
Different from the above, we propose
temporal-aware prompting with foundation segmentation models (\eg, SAM) to tackle RVOS problems.
\section{Proposed Method}


\begin{figure*}[t!]
	\centering
	\includegraphics[width=1.0\linewidth]{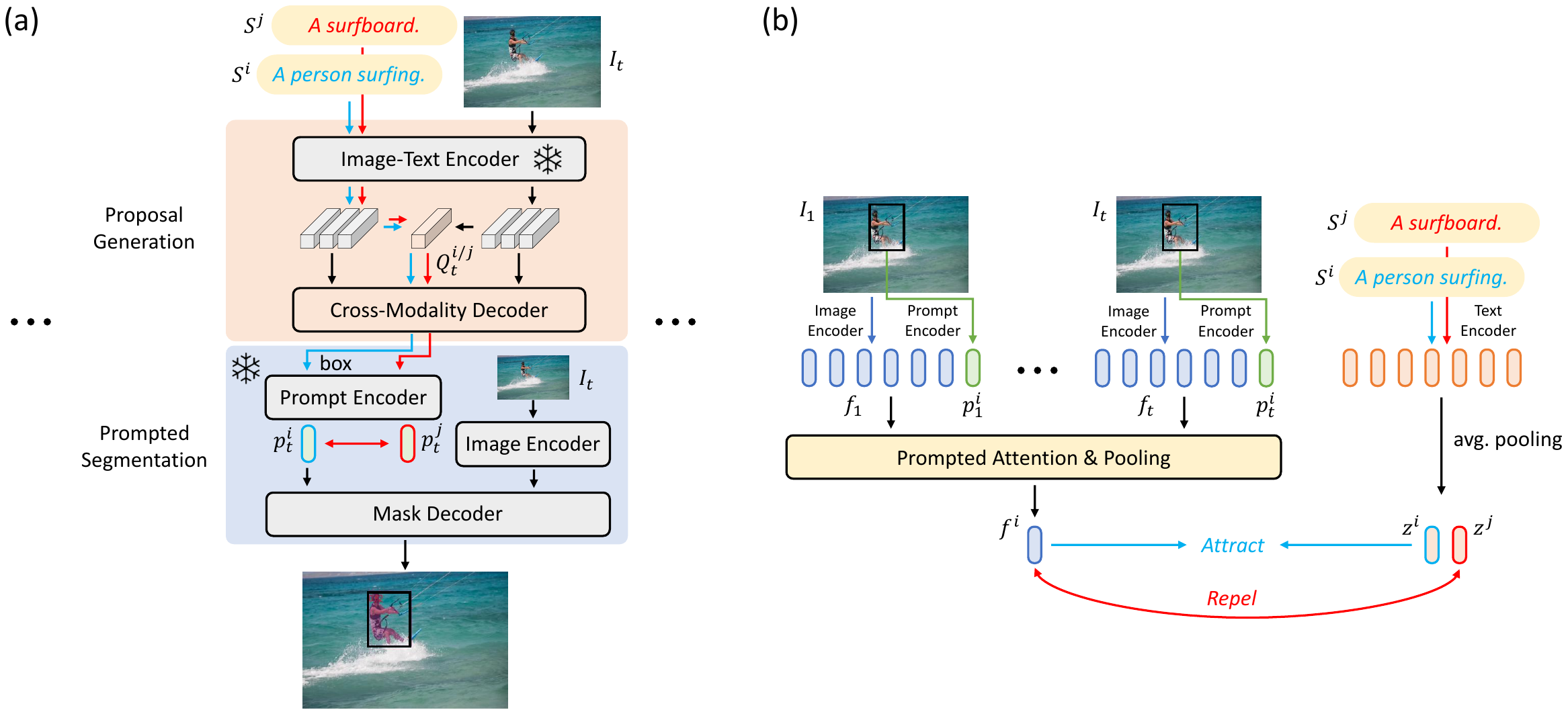}

    \caption{Overview of our proposed \nameshort~framework. In (a), our proposal generation takes each frame $I_t$ and the referring sentence $S^i$ to derive object queries $Q_t^i$ and produce the prompt embedding $p_t^i$ for segmentation, with another sentence $S^j$ as input for performing \textit{\namea}. In (b), to handle sentence descriptions containing long-term motions or actions in referring video object segmentation, we uniquely present \textit{\nameb} to align the text with the referred object at the video level.}

	\label{fig:model}
\end{figure*}

\subsection{Overview}

\paragraph{Problem Definition.} For the sake of completeness, we first define the problem setting and notations used in this paper. In Referring Video Object Segmentation (RVOS), we assume that the training data contain a set of $N$ videos, where each video $V=\{I_t\}_{t=1}^T$ is a sequence of $T$ frames and is associated with a set of referring sentences $S=\{S^i\}_{i=1}^M$ describing $M$ distinct objects. The goal of RVOS is to produce segmentation masks for the referred objects. Different from previous works~\cite{wu2022language, wu2023onlinerefer, miao2023spectrum} which require dense mask annotations for training, we assume that we only have access to box-level annotations $\hat{B}^i=\{\hat{B}^i_t\}_{t=1}^T$ for the $T$ frames corresponding to the $i$th referring sentence $S^i$, where each bounding box $\hat{B}_t^i = (\hat{x}_t^i, \hat{y}_t^i, \hat{h}_t^i, \hat{w}_t^i)$ is represented by the coordinate of the center point and the height and width.

\paragraph{Framework Overview.} Under the above setting, our goal is to efficiently adapt image-based foundation segmentation models for addressing referring video object segmentation from such weak supervision. To achieve efficient model adaptation, we propose a novel \textit{\namebf~(\textbf{\nameshort})} framework, which advances vision-language learning to produce temporal-consistent yet text-aware position prompts for segmentation purposes. As shown in Figure~\ref{fig:model}, our proposed \nameshort~framework is designed to generate the bounding box proposal by taking object queries to perform cross-modal attention at each frame. Such proposals then serve as position prompts to instruct foundation segmentation models to segment the referred object. 
To facilitate the position prompts to be text- and temporal-aware, we propose \textit{\namefbf~(\textbf{\namefabbr}), including: 1)}
\textit{\namea~(\nameaabbr)}~at the frame level, which encourages the output proposals to be distinct when taking different referring sentences as input; 2)
\textit{\nameb~(\namebabbr)}, which aims to align the output proposal sequence and its corresponding object with the input text for each video clip. 
With the proposed \namefabbr, our \nameshort~framework would produce temporal-consistent yet text-aware position prompts for the referred object, enabling efficient adaptation from weak supervision without additional finetuning for foundation models.

\subsection{\name}

Recent foundation segmentation models{~\cite{kirillov2023segment,wang2023seggpt,zou2023segment}} have presented overwhelming performance on various segmentation tasks. When prompted by points or bounding boxes indicating the positions, these foundation models would produce high-quality object masks as desired. However, existing foundation segmentation models are mainly trained from general image data and therefore have limited ability to comprehend video content or complex text descriptions. To adapt image-based foundation segmentation models to address referring video object segmentation, our proposed \nameshort~framework is designed to learn and generate position prompts for the target object from the input video frames and the referring sentences. In this way, our \nameshort~framework enables efficient model adaptation without additional finetuning for foundation models, avoiding possible overfitting issues while reducing computational cost and time. We now detail our learning scheme below.

\subsubsection{Weakly-Supervised Position Prompts}

To produce precise position prompts for segmentation, we advance vision-language learning to generate bounding box proposals for the referred object. As illustrated in Figure~\ref{fig:model}, our \nameshort~framework first employs a Transformer-based image-text encoder to extract visual features and linguistic features for each frame $I_t$ and the referring sentence $S^i$, respectively. Inspired by~\cite{liu2023grounding}, we adopt the query generation mechanism to obtain a set of object queries $Q_t^i$. By taking visual features and linguistic features as keys and values, the derived object queries $Q_t^i$ would perform cross-attention through the cross-modality decoder to generate the box proposal $B_t^i$. With the ground-truth bounding box $\hat{B}_t^i$, the standard box loss $L_{box}$ is formulated by the regression loss and generalized IoU loss $L_{g}$~\cite{rezatofighi2019generalized}:
\begin{equation}
\begin{aligned}
\label{eq:loss_box}
L_{box} = \mathbb{E}_{V, S^i}\left[\sum_{i=1}^T \lambda_{r}\| B_t^i - \hat{B}_t^i \|_1 + \lambda_{g}L_{g}(B_t^i, \hat{B}_t^i)\right]
\end{aligned}
\end{equation}
where $\lambda_{r}$ and $\lambda_{g}$ are hyper-parameters for the two loss terms, respectively. Here, since there is typically only one target object in referring segmentation tasks, we simply select the output proposal $B_t^i$ with the highest confidence score at each frame instead of using the Hungarian loss~\cite{carion2020end} for matching. It is worth noting that we do not need mask loss for training like most existing RVOS works~\cite{wu2022language,han2023html,wu2023onlinerefer,miao2023spectrum,tang2023temporal,luo2023soc,wu2023segment}.



\subsubsection{\namef}
In referring segmentation tasks, the sentence descriptions could be ambiguous. For example, the sentence ``A person surfing'' in Figure~\ref{fig:model} refers to the person alone rather than both the person and the surfboard. To mitigate such text ambiguity in natural language, we propose to perform \textit{\namea~(\nameaabbr)} at the frame level to generate distinct proposals for different referring sentences. 
Apart from the text ambiguity, the sentence descriptions in referring video object segmentation often contain long-term motions or actions. Sentences like ``a gold fish on the left swimming towards the top right'' require considering all the frames as a whole to perform video segmentation. To align the text with the referred object at the video level, we uniquely present \textit{\nameb~(\namebabbr)}. 
The learning scheme is detailed below.

\paragraph{\namea.}

\begin{table*}[t!]
    \setlength{\tabcolsep}{8pt}
    \centering
    \small
	\resizebox{1.01\textwidth}{!}{
    
    \begin{tabular}{l | c | c | c c c | c c c}
    \toprule 
    \multirow{2}{*}{Method} & \multirow{2}{*}{Publication} & \multirow{2}{*}{Referring \& Video Training Data} & \multicolumn{3}{c |}{Ref-YouTube-VOS} & \multicolumn{3}{c}{Ref-DAVIS17} \\
     & &  & \( \mathcal{J} \)\&\( \mathcal{F} \) & \( \mathcal{J} \) & \( \mathcal{F} \) &  \( \mathcal{J} \)\&\( \mathcal{F} \) & \( \mathcal{J} \) & \( \mathcal{F} \) \\
    \midrule 
    URVOS~\cite{seo2020urvos} & ECCV'20 & RefYT & 47.2 & 45.3 & 49.2 & 51.5 & 47.3 & 56.0 \\
    

    MTTR~\cite{MTTR} & CVPR'22 & RefYT & 55.3 & 54.0 & 56.6 & - & - & -\\ 
    ReferFormer~\cite{wu2022language} & CVPR'22 & RefC, RefYT & 62.9 & 61.3 & 64.6 & 61.1 & 58.1 & 64.1 \\
    MANet~\cite{MANet} & ACM MM'22 & RefYT & 55.6 & 54.8 & 56.5 & - & - & - \\
    LOCATER~\cite{liang2023local} & TPAMI'23 & RefYT & 56.5 & 54.8 & 58.1 & - & - & - \\ 
    VLT~\cite{ding2022vlt} & TPAMI'23 & RefC, RefYT & 63.8 & 61.9 & 65.6 & 61.6 & 58.9 & 64.3 \\
    $\text{R}^2\text{-VOS}$~\cite{li2023robust} & ICCV'23 & RefC, RefYT & 61.3 & 59.6 & 63.1 & - & - & - \\
    HTML~\cite{han2023html} & ICCV'23 & RefC, RefYT & 63.4 & 61.5 & 65.2 & 62.1 & 59.2 & 65.1 \\
    OnlineRefer~\cite{wu2023onlinerefer} & ICCV'23 & RefC, RefYT & 63.5	& 61.6 & 65.5 & 64.8 & 61.6	& 67.7 \\    
    SgMg~\cite{miao2023spectrum}  & ICCV'23 & RefC, RefYT & 65.7 & 63.9 & 67.4 & 63.3 & 60.6 & 66.0 \\ 
    TempCD~\cite{tang2023temporal} & ICCV'23 & RefC, RefYT & 65.8 & 63.6 & 68.0 & 64.6 & 61.6 & 67.6 \\    
    SOC~\cite{luo2023soc} & NeurIPS'23 & RefC, RefYT & 67.3 & 65.3 & 69.3 & 65.8 & 62.5 & 69.1 \\
    LoSh~\cite{yuan2023losh} & arXiv'23 & RefC, RefYT & 64.2 & 62.5 & 66.0 & 62.5 & 59.5 & 65.4 \\
    RefSAM~\cite{li2023refsam} & arXiv'23 & RefC, RefYT & 62.1 & 60.9 & 63.3 & 69.5 & 65.9 & 73.2 \\
    EPCFormer~\cite{chen2023epcformer} & arXiv'23 & RefYT, AVOS & 65.0 & 62.9 & 67.2 & - & - & - \\    
    \midrule


    UniNEXT~\cite{yan2023universal} & CVPR'23 & RefC, RefYT, G, La, T, YT, B, V, O & 66.2 & 64.0 & 68.4 & 66.7 & 62.3 & 71.1 \\
    DEVA~\cite{cheng2023tracking} & ICCV'23 & RefC, RefYT, YT, D, O & 66.0 & - & - & 66.3 & - & - \\
    UniRef~\cite{wu2023segment} & ICCV'23 & RefC, RefYT, RefD, YT, O, LV & 67.4 & 65.5 & 69.2 & 66.3 & 62.9 & 69.7 \\ 
    MUTR~\cite{yan2023referred} & arXiv'23 & RefC, RefYT, AVSB & 68.4 & 66.4 & 70.4 & 68.0 & 64.8 & 71.3 \\ \midrule
    WRVOS~\cite{zhao2023learning} & arXiv'23 & RefYT (box + 1st-frame mask) & 46.6 & 45.6 & 47.6 & 47.3 & 44.6 & 50.0 \\
    Grounded-SAM~\cite{liu2023grounding} & arXiv'23 & RefC (box) & 62.3 & 61.0 & 63.6 & 65.2 & 62.3 & 68.0 \\
    \nameshort~(\textbf{Ours}) & - & RefC (box), RefYT (box) & 65.5 & 64.1 & 66.9 & 70.6 & 67.8 & 73.3 \\
    \bottomrule 
    \end{tabular}
    }
    \caption{Quantitative comparison to state-of-the-art methods on the validation split of Ref-YouTube-VOS and Ref-DAVIS17. RefYT: Ref-YouTube-VOS, RefD: Ref-DAVIS, RefC: RefCOCO~\cite{mao2016generation,yu2016modeling}, AVOS: Audio-VOS~\cite{pan2022wnet}, AVSB: AVSBench~\cite{zhou2022audio}, YT: YouTube-VOS 2019~\cite{xu2018youtube}, D: DAVIS17~\cite{perazzi2016benchmark}, O: Occluded VIS~\cite{qi2022occluded}, LV: Long-term VOS~\cite{hong2023lvos}, G: GOT-10K~\cite{huang2019got}, La: LaSOT~\cite{fan2019lasot}, T: TrackingNet~\cite{muller2018trackingnet}, B: BDD100K~\cite{yu2020bdd100k}, V: VIS19~\cite{yang2019video}.
    } \label{tab:quantitative}
\end{table*}

Formally, in addition to the input sentence $S_i$, we forward another sentence $S^j$ through our \nameshort~framework to obtain the output proposal $B_t^j$ for another object at each frame. To perform contrastive learning, we leverage the prompt encoder from the foundation segmentation models to extract the prompt embeddings $p_t^i$, $p_t^j$, and $\hat{p}_t^i$ for the proposals $B_t^i$ and $B_t^j$ and the ground-truth bounding box $\hat{B}_t^i$, respectively. By taking $p_t^i$, $\hat{p}_t^i$, and $p_t^j$ as the anchor, positive, and negative sample, the frame-level triplet contrastive loss $L_{contra}^f$ would be computed as follows:
\begin{equation}
\begin{aligned}
&L_{contra}^f = \mathbb{E}_{V, S^i, S^j} \left[ \sum_{t=1}^{T} \max(0, d_t^{p} - d_t^{n}) \right],\\
&\text{where} \quad d_t^{p} = \|p_t^i - \hat{p}_t^i\|_2  \quad \text{and} \quad d_t^{n} = \|p_t^i - p_t^j\|_2.
\end{aligned}
\end{equation}
We note that to preserve the latent space learned by foundation models for segmentation, we choose to freeze the prompt encoder during training. Under the guidance of the prompt encoder, our proposed \textit{\nameaabbr} enforces the distinctness of the proposals while enhancing the position prompts to be text-aware.

\paragraph{\nameb.}

In addition to the prompt embedding $p_t^i$ derived in \textit{\namea}, we also utilize the image encoder to extract the visual features $f_t$. With the cross-attention performed at each frame by taking the prompt embedding $p_t^i$ as the query and visual features $f_t$ as keys and values, followed by an average pooling layer for temporal aggregation, the video-level content feature $f^i$ would be encoded for the referred object. As for the referring sentences $S^i$ and $S^j$, we derive the sentence-level linguistic features $z^i$ and $z^j$ from the text encoder. Then, the video-level triplet contrastive loss $L_{contra}^v$ would be computed as follows:
\begin{equation}
\begin{aligned}
&L_{contra}^v= \mathbb{E}_{V, S^i, S^j} \left[ \max(0, d^{p} - d^{n}) \right],\\
&\text{where} \quad d^{p} = \|f^i - z^i\|_2  \quad \text{and} \quad d^{n} = \|f^i - z^j\|_2.
\end{aligned}
\end{equation}
Note that the prompt, image, and text encoders are all frozen during training to preserve their pretrained semantic spaces while avoiding overfitting.

Finally, we define the total loss function $L_{total}$ as:
\begin{equation}
\begin{aligned}
&L_{total}= L_{box} + L_{contra},
\end{aligned}
\end{equation}
where $L_{contra}= \lambda_{f}L_{contra}^f + \lambda_{v}L_{contra}^v$, and $\lambda_{f}$ and $\lambda_{v}$ are hyper-parameters for the two contrastive loss, respectively. 
With the proposed TAP-CL, 
our \nameshort~framework would produce temporal-consistent yet text-aware bounding box proposals, allowing video segmentation by taking the learned proposals to prompt image-based foundation segmentation models. It is worth repeating that, the above learning scheme does not require any dense mask annotations. Furthermore, our proposed \nameshort~framework learns to prompt instead of finetuning foundation models, enabling efficient adaptation to referring video object segmentation from weak supervision.

\section{Experiments}

\begin{table*}[t]
	\centering
	\small
	\resizebox{1\textwidth}{!}{
		\setlength\tabcolsep{8pt}
		\renewcommand\arraystretch{1.0}
		\begin{tabular}{l|c|ccccc|cc}
			\toprule
			Method & Publication & P@0.5 & P@0.6 & P@0.7 & P@0.8 & P@0.9 & Overall IoU & Mean IoU\\ 
			\midrule
			CMSA + CFSA~\cite{ye2021referring} & TPAMI'21 & 48.7& 43.1 & 35.8& 23.1 &5.2& 61.8& 43.2 \\
			MTTR~\cite{MTTR} & CVPR'22 & 75.4 & 71.2 & 63.8 & 48.5 & 16.9 & 72.0 & 64.0 \\
   			ReferFormer~\cite{wu2022language} & CVPR'22 &  83.1 & 80.4 & 74.1 & 57.9 & 21.2 & 78.6 & 70.3 \\ 
               LOCATER~\cite{liang2023local} & TPAMI'23 & 70.9 & 64.0 & 52.5 & 35.1 & 10.1 & 69.0 & 59.7 \\ 
            TempCD~\cite{tang2023temporal} & ICCV'23 & - & - & - & - & - & 76.6 & 68.6 \\
            HTML~\cite{han2023html} & ICCV'23 & 84.0 & 81.5 & 75.8 & 59.2 & 22.8 & 79.5 & 71.2 \\
		OnlineRefer~\cite{wu2023onlinerefer} & ICCV'23 & 83.1 & 80.2	& 73.4	& 56.8	&21.7	& 79.6 & 70.5\\
              LoSh~\cite{yuan2023losh} & arXiv'23 & 85.3 & 80.7 & 74.3 & 57.7 & 21.1 & 78.9 & 71.3 \\
            WRVOS~\cite{zhao2023learning} & arXiv'23 & 62.9 & 58.0 & 49.8 & 35.0 & 11.0 & 66.3 & 53.9 \\
            Grounded-SAM~\cite{liu2023grounding} & arXiv'23 & 80.8  & 78.7  & 73.9  & 59.8  & 23.8  & 71.0  & 68.9 \\
            \nameshort~(\textbf{Ours}) & - & 83.9  & 81.5  & 75.9  & 60.5  & 23.4  & 77.3  & 71.3 \\
			\bottomrule
		\end{tabular}
	}
        \vspace{-2mm}
	\caption{\textbf{The quantitative evaluation on A2D-Sentences}, with Precision@K, Overall IoU and Mean IoU.}
	\vspace{1mm}
	\label{table:sota_a2d}
\end{table*}

\begin{table*}[t]
	\centering
	\small
	\resizebox{1\textwidth}{!}{
		\setlength\tabcolsep{8pt}
		\renewcommand\arraystretch{1.0}
		\begin{tabular}{l|c|ccccc|cc}
			\toprule
			Method & Publication & P@0.5 & P@0.6 & P@0.7 & P@0.8 & P@0.9 & Overall IoU & Mean IoU\\ 
			\midrule
			CMSA + CFSA~\cite{ye2021referring} & TPAMI'21 & 76.4 & 62.5 & 38.9 & 9.0 & 0.1 & 62.8 & 58.1 \\
			MTTR~\cite{MTTR} & CVPR'22 & 93.9 & 85.2 & 61.6 & 16.6 & 0.1 & 70.1 & 69.8 \\
		      ReferFormer~\cite{wu2022language} & CVPR'22 & 96.2 &90.2 & 70.2 & 21.0 & 0.3 & 73.0 & 71.8 \\ 
                LOCATER\cite{liang2023local} & TPAMI'23 & 89.3 & 77.2 & 50.8 & 10.6 & 0.2 & 67.3 & 66.3 \\
                TempCD\cite{tang2023temporal} & ICCV'23 & - & - & - & - & - & 70.6 & 69.6\\
		      OnlineRefer~\cite{wu2023onlinerefer} & ICCV'23 & 96.1	& 90.4 &	71.0	& 21.9 &	0.2	& 73.5	& 71.9\\
                SgMg~\cite{miao2023spectrum} & ICCV'23 & - & - & - & - & - & 73.7 & 72.5 \\
                SOC~\cite{luo2023soc} & NeurIPS'23 & 96.9 & 91.4 & 71.1 & 21.3 & 0.1 & 73.6 & 72.3\\
                LoSh~\cite{yuan2023losh} & arXiv'23 & 96.3 & 90.1 & 70.4 & 20.5 & 0.2 & 73.2 & 72.5 \\            
                EPCFormer~\cite{chen2023epcformer} & arXiv'23 & 97.6 & 93.1 & 72.6 & 23.0 & 0.0 & 74.0 & 73.1 \\
                WRVOS~\cite{zhao2023learning} & arXiv'23 & 82.0 & 67.3 & 41.4 & 8.9 & 0.1 & 63.2 & 62.7 \\            
                Grounded-SAM~\cite{liu2023grounding} & arXiv'23 & 95.3  & 89.4  & 70.0  & 23.1  & 0.3  & 71.9  & 71.7 \\
                \nameshort~(\textbf{Ours}) & - & 96.8  &90.8  & 71.0  & 23.2  &  0.3 & 73.3  & 72.4 \\

			\bottomrule
		\end{tabular}
	}
        \vspace{-2mm}
	\caption{\textbf{The quantitative evaluation on JHMDB-Sentences}, with Precision@K, Overall IoU and Mean IoU.}
	\vspace{-3mm}
	\label{table:sota_jhmdb}
\end{table*}

\begin{figure*}[t!]
	\centering
	\includegraphics[width=1.0\linewidth]{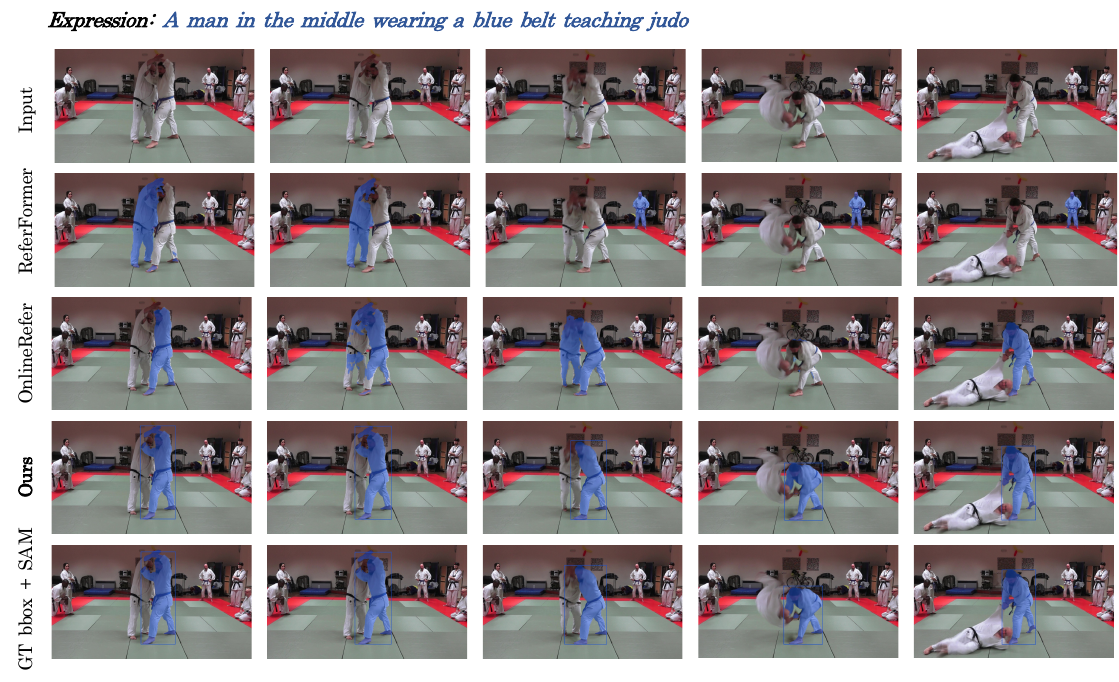}
    \vspace{-7mm}
    \caption{Qualitative comparisons of the state-of-the-art methods on Refer-DAVIS$_{17}$, where ``GT-bbox + SAM'' represents the result by taking ground-truth bounding boxes to prompt SAM.}
    \vspace{2mm}
	\label{fig:vis_davis}
\end{figure*}

\begin{figure*}[t!]
	\centering
	\includegraphics[width=1.0\linewidth]{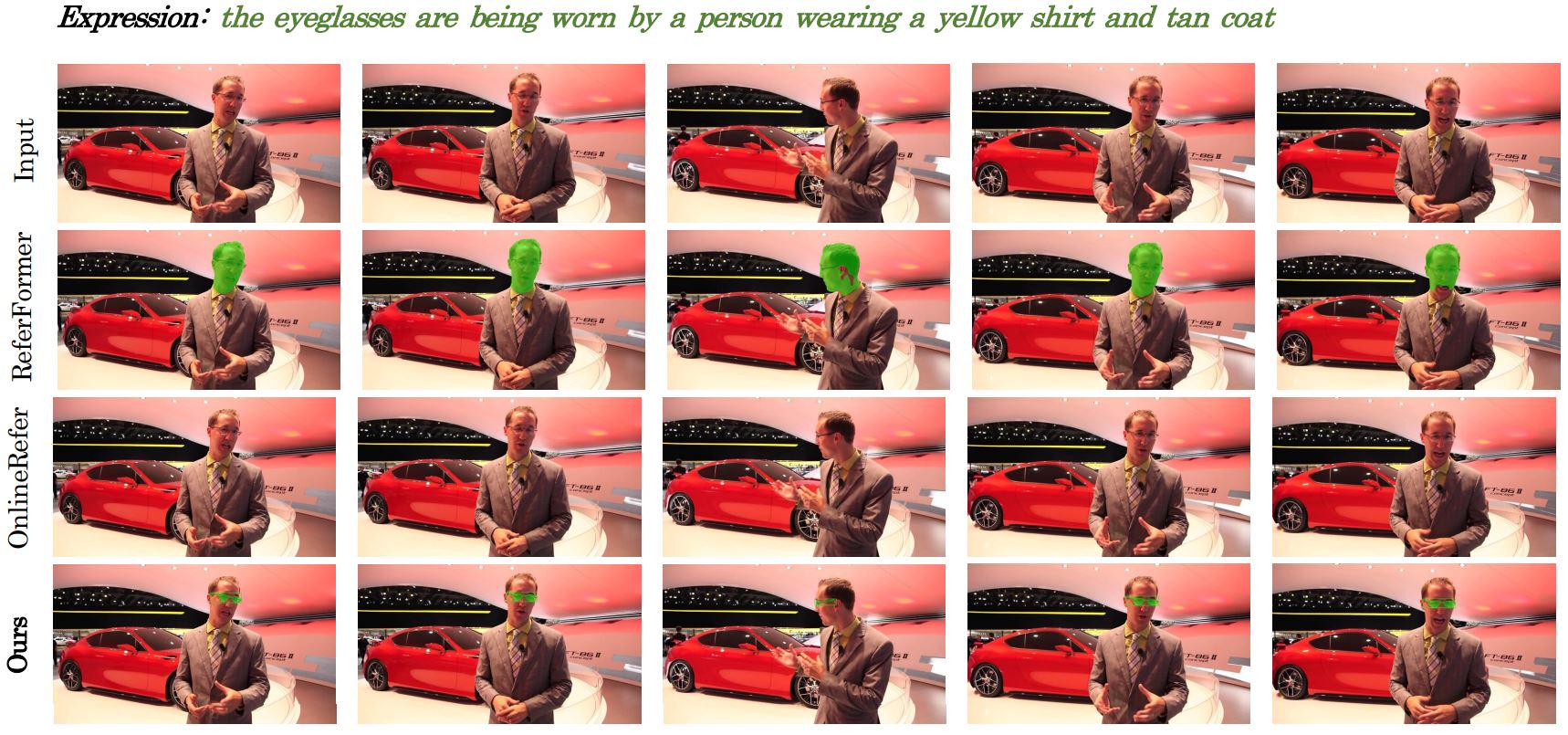}

    \caption{Qualitative comparisons of the state-of-the-art methods on Refer-Youtube-VOS.}

	\label{fig:vis_ytvos}
\end{figure*}

\subsection{Datasets and Evaluation Metrics}

\paragraph{Datasets.} We conduct experiments on four RVOS benchmark datasets: Refer-Youtube-VOS~\cite{seo2020urvos}, Refer-DAVIS$_{17}$~\cite{khoreva2018video}, A2D Sentences~\cite{gavrilyuk2018actor}, and J-HMDB Sentences~\cite{gavrilyuk2018actor}. Refer-Youtube-VOS is a large-scale dataset for RVOS, with $3,975$ videos, $7,451$ objects, and $27,899$ expressions. Refer-DAVIS$_{17}$ is augmented from the popular video object segmentation dataset, DAVIS$_{17}$~\cite{caelles20182018}. It contains $90$ videos ($60$ for training and $30$ for testing) with more than $1,500$ expressions. A2D Sentences~\cite{gavrilyuk2018actor} and J-HMDB Sentences~\cite{gavrilyuk2018actor} are extended from the A2D~\cite{xu2015can} and J-HMDB~\cite{jhuang2013towards} datasets with sentences describing the actors and actions appearing in the video content. A2D Sentences contains $3,036$ training videos
and $746$ testing videos with a total of $6,656$ sentences, while J-HMDB Sentences contains $928$ video clips of $21$ different actions and $928$ sentences.

\paragraph{Evaluation Metrics.} For the Ref-Youtube-VOS and Ref-DAVIS$_{17}$ datasets, we follow the standard protocol and adopt the following evaluation metrics: region similarity $\mathcal{J}$ (average IoU), contour accuracy $\mathcal{F}$ (average boundary similarity), and their mean value $\mathcal{J}\&\mathcal{F}$. Since the annotations of the Ref-Youtube-VOS validation set are not publicly released, we evaluate the results on the official server. As for Ref-DAVIS$_{17}$, we use the official code for evaluation. For A2D Sentences and J-HMDB Sentences, we adopt Precision@K, Overall IoU, and Mean IoU for evaluation. Overall IoU is the ratio between the total intersection and the union area over all the testing data, and Mean IoU is the averaged IoU over the testing data. Precision@K measures the
percentage of testing data with IoU score higher than a threshold K, where K $\in \left[0.5, 0.6, 0.7, 0.8, 0.9\right]$.

\subsection{Implementation Details}
\vspace{-1mm}
We follow from~\cite{wu2023onlinerefer, wu2022language} to train our model on the Ref-YouTube-VOS dataset, and directly evaluate by the validation set provided by Ref-YouTube-VOS and Ref-DAVIS$_{17}$. For our detailed model architecture,  our image-text encoder comprises Swin-Transformer~\cite{liu2021swin} for the image features and BERT~\cite{devlin2018bert} for the text features. Besides, we set up our cross-modality decoder with 6 cross-attention transformer layers. For the segmentation part, we take SAM as our main segmentor to take our special text-aware position prompt as input. Thus, the prompt encoder, image encoder, and mask decoder are followed by SAM in our setting. We set the learning rate to $0.0001$ and train our framework for $12$ epochs. Following~\cite{liu2023grounding}, we set $\lambda_r$ and $\lambda_g$ as $5$ and $2$ respectively. As for $\lambda_f$ and $\lambda_v$, we use $0.01$ and $0.1$ on Ref-Youtube-VOS and $0.0001$ and $0.001$ on A2D Sentences, respectively. We implement our framework in PyTorch and train the model on $8$ NVIDIA V100 GPUs.



\subsection{Quantitative and Qualitative Comparisons}
\vspace{-2mm}
To evaluate our proposed \nameshort~framework, we first provide quantitative comparisons with state-of-the-art methods on Refer-Youtube-VOS~\cite{seo2020urvos} and Refer-DAVIS$_{17}$~\cite{khoreva2018video}. As shown in Table~\ref{tab:quantitative}, we see that our \nameshort~framework achieves $65.5\%$ and $70.6\%$ in $\mathcal{J}\&\mathcal{F}$ on Refer-Youtube-VOS and Refer-DAVIS$_{17}$, respectively. Compared with RefSAM~\cite{chen2023epcformer}, our \nameshort~framework is $3.4\%$ and $1.1\%$ higher on the two datasets. This validates that our learned position prompts would properly instruct foundation segmentation models to perform referring video object segmentation. While UniRef~\cite{wu2023segment} and MUTR~\cite{yan2023referred} achieve competitive performance on Refer-Youtube-VOS, these methods require large-scale referring or video data for training. Compared to WRVOS~\cite{zhao2023learning}, which observes box-level supervision plus the mask annotation for the first frame, our \nameshort~framework is over $20\%$ higher with box-level supervision only. Similar results are observed on A2D Sentences~\cite{gavrilyuk2018actor} and J-HMDB Sentences~\cite{gavrilyuk2018actor}. In Table~\ref{table:sota_a2d}, our method reports $71.3\%$ in Mean IoU. As for J-HMDB Sentences, we achieve $72.4\%$ in Table~\ref{table:sota_jhmdb}.

In Figure~\ref{fig:vis_davis} and~\ref{fig:vis_ytvos}, we also provide qualitative comparisons with ReferFormer~\cite{wu2022language} and OnlineRefer~\cite{wu2023onlinerefer} on Refer-DAVIS$_{17}$ and Refer-Youtube-VOS~\cite{seo2020urvos}. We observe that, our method outperforms OnlineRefer and the produced bounding box proposals are close to the ground-truth bounding boxes. From the above experiments, we validate that our proposed \nameshort~framework would produce position prompts from weak supervision, enabling efficient adaptation of image-based foundation segmentation models for addressing referring video object segmentation.

\begin{table}[!t]
	\centering
	\resizebox{1.0\linewidth}{!}{
		\begin{tabular}{l c c c c}
			\toprule
			Method & Weak sup. & Online & Decoupled & Addi. Training Videos \\
			\midrule
			ReferFormer~\cite{wu2022language}    &                &                &            & No Need \\
                WRVOS~\cite{zhao2023learning}        &    \checkmark            &      &            & No Need \\
                OnlineRefer~\cite{wu2023onlinerefer} &      &   \checkmark             &            & No Need \\
                
			DEVA~\cite{cheng2023tracking}        &                &                & \checkmark & YT, D, O \\
			\nameshort~(\textbf{Ours})            &\checkmark      & \checkmark     & \checkmark & No Need \\
			
			\bottomrule
	\end{tabular}}
         \caption{Setting comparisons with recent RVOS methods. ``Weak sup.'': Trained mainly with box-level weak supervisions, ``Online'': Online method rather than offline method,  ``Decoupled'': Decoupled segmentation instead of end-to-end training, ``Addi. Training Videos'': Additional video datasets for training. YT: YouTube-VOS 2019~\cite{xu2018youtube}, D: DAVIS17~\cite{perazzi2016benchmark}, O: Occluded VIS~\cite{qi2022occluded}.}
         \vspace{-2mm}
	\label{tab:setting}
\end{table}

                
			
\subsection{Setting and Efficiency Comparisons}
			

\begin{table}[!t]
	\centering
	\resizebox{1.0\linewidth}{!}{
		\begin{tabular}{l c c c }
			\toprule
			Method &  \# of trainable parameters & Ref-YTVOS & Ref-DAVIS \\
			\midrule
			ReferFormer~\cite{wu2022language}    &     $\sim$112M   & 62.9 & 61.1   \\
                OnlineRefer~\cite{wu2023onlinerefer} &     $\sim$221M  & 63.5 & 64.8 \\
			DEVA~\cite{cheng2023tracking}        &     $\sim$112M  & 66.0 & 66.3 \\
			\nameshort~(\textbf{Ours})           &     $\sim$15M & 65.5 & 70.6 \\
			
			\bottomrule
	\end{tabular}}
         \caption{
         Efficiency comparisons with recent RVOS methods, along with the \( \mathcal{J} \)\&\( \mathcal{F} \) scores on Ref-YouTube-VOS and Ref-DAVIS17. 
         }
	\label{tab:parameter}
        \vspace{-5mm}
\end{table}

In Table~\ref{tab:setting}, we compare the setting of our proposed \nameshort~framework with recent RVOS methods. From this table, we see that WRVOS~\cite{zhao2023learning} attempts to address RVOS from box-level weak supervision plus the ground-truth mask for the first frame, while OnlineRefer~\cite{wu2023onlinerefer} extends ReferFormer~\cite{wu2022language} with query propagation to handle ongoing videos under the online setting. However, these methods require end-to-end training for vision-language models, which could be computationally expensive and time-consuming. On the other hand, assuming that additional video data are accessible, DEVA~\cite{cheng2023tracking} decouples RVOS into image segmentation and temporal propagation to increase the scalability. Compared to these works, our proposed \nameshort~framework decouples RVOS into proposal generation and prompted segmentation with no need for additional video data for training. In this decoupled manner, our framework can learn proper prompts from weak supervision for foundation segmentation models and could also be applied to online settings. 

In Table~\ref{tab:parameter}, 
we also provide efficiency comparisons with recent works. We see that the number of trainable parameters of our method is over $7$ times fewer than DEVA. This is because that our proposed \nameshort~framework learns to prompt foundation models for efficient adaptation instead of training a vision-language model end-to-end. Together with the quantitative comparisons in Table~\ref{tab:quantitative}, we validate that our proposed \nameshort~framework is preferable in terms of performance, setting, and efficiency.

\subsection{Ablation Studies}
\vspace{-1mm}
To verify the effectiveness of our proposed loss functions, we conduct ablation studies by taking the ground-truth bounding boxes to compute the IoU scores of the predicted box proposals on Ref-DAVIS17. From Table~\ref{tab:ablation}, we see that when only $L_{box}$ is considered, the box and segmentation score \( \mathcal{J} \)\&\( \mathcal{F} \) would improve $3.3\%$ and $4.6\%$ compared to Grounded-SAM~\cite{liu2023grounding}. If we further apply our proposed $L_{contra}$ to perform contrastive learning at frame level and video level, the box and segmentation score would improve to $74.4\%$ and $70.6\%$, which are $1.2\%$ and $0.8\%$ higher. Finally, if we directly take the ground-truth boxes to prompt SAM, the superior performance of $83.6\%$ in \( \mathcal{J} \)\&\( \mathcal{F} \) would be observed. This demonstrates that image segmentation could be mostly solved by SAM, and therefore how to generate proper prompts to instruct foundation segmentation models for referring segmentation tasks would now be of interest. From the above experiments, we confirm that our proposed loss functions would learn precise position prompts (box proposals) from the referring sentence and the input video, allowing efficient adaptation of foundation models for addressing RVOS.
\begin{table}[!t]
	\centering
	\resizebox{1.0\linewidth}{!}{
		\begin{tabular}{l | c | c c c}
			\toprule
			Method & Box & \( \mathcal{J} \)\&\( \mathcal{F} \) & \( \mathcal{J} \) & \( \mathcal{F} \) \\
			\midrule
			Grounded-SAM~\cite{liu2023grounding}  & 69.9  & 65.2 & 62.3 & 68.0 \\
                \nameshort~($L_{box}$ only)           & 73.2  & 69.8 & 67.0 & 72.7 \\
                \nameshort~($L_{box}$ + $L_{contra}$) & 74.4  & 70.6 & 67.8 & 73.3 \\
                \midrule
                GT Box + SAM (upper bound)           & 100.0  & 83.6 & 80.1 & 87.2 \\
			
			\bottomrule
	\end{tabular}}
         \caption{Ablation studies of the loss functions on Ref-DAVIS17. ``Box'': IoU scores of the bounding boxes (position prompts).}
         \vspace{-5mm}
	\label{tab:ablation}
\end{table}

\vspace{-2mm}

\section{Conclusion}
\label{sec:conclusion}
\vspace{-2mm}
In this work, we propose the \namebf~(\textbf{\nameshort}) framework to efficiently adapt foundation segmentation models for addressing RVOS from weak supervision. 
More specifically, we propose \textit{\namefbf~(\textbf{\namefabbr})} to enhance the association between the position prompts and the referring sentences with only box supervisions, including \textit{\namea~(\nameaabbr)}~and \textit{\nameb~(\namebabbr)} at frame level and video level, respectively. 
With the proposed \namefabbr, our \nameshort~framework can generate temporal-consistent yet text-aware position prompts describing locations and movements for the referred object from the video. With no need of additional finetuning for foundation segmentation models, we are able to produce precise masks for the referred object in the video.
The experimental results in the standard RVOS benchmarks (Ref-YouTube-VOS, Ref-DAVIS17, A2D-Sentences, and JHMDB-Sentences) demonstrate the competitive performance of our proposed \nameshort~framework given only bounding box weak supervision.



{
    \small
    \bibliographystyle{ieeenat_fullname}
    \bibliography{main}
}



\end{document}